\pdfoutput=1
\documentclass[11pt]{article}

\usepackage[review]{acl}

\usepackage{times}
\usepackage{latexsym}

\usepackage{amsmath,amsthm}
\newtheorem{definition}{Definition}
\newtheorem{conjecture}{Conjecture}
\usepackage{amsmath,amsfonts,bm}
\usepackage{amsthm}

\def\vv{{\bm{v}}}

\def\vx{{\bm{x}}}
\def\vy{{\bm{y}}}

\usepackage{tikz}%
\usetikzlibrary{positioning}
\usepackage[T1]{fontenc}
\usepackage[utf8]{inputenc}

\usepackage{microtype}
\usepackage{hyperref}
\hypersetup{
    colorlinks=true,
    urlcolor=magenta,
}
\usepackage{inconsolata}

\usepackage{graphicx}
\usepackage{float}
\usepackage{subcaption}
\usepackage{booktabs}

\newcommand{\R}{\mathbb{R}}

\title{From Text to Trajectories: GPT-2 as an ODE Solver via In-Context}

\author{\textbf{Ziyang Ma}\textsuperscript{1}, \textbf{Baojian Zhou}\textsuperscript{2}, \textbf{Deqing Yang}\textsuperscript{2} ,  \textbf{Yanghua Xiao}\textsuperscript{1}\\
\textsuperscript{1} College of Computer Science and Artificial Intelligence, Fudan University\\
\textsuperscript{2} the School of Data Science, Fudan University\\
\texttt{zyma25@m.fudan.edu.cn},\\
\texttt{\{bjzhou,yangdeqing,shawyh\}@fudan.edu.cn}
}

\begin{document}
\maketitle
\begin{abstract}
In-Context Learning (ICL) has emerged as a new paradigm in large language models (LLMs), enabling them to perform novel tasks by conditioning on a few examples embedded in the prompt. Yet, the highly nonlinear behavior of ICL for NLP tasks remains poorly understood. To shed light on its underlying mechanisms, this paper investigates whether LLMs can solve ordinary differential equations (ODEs) under the ICL setting. We formulate standard ODE problems and their solutions as sequential prompts and evaluate GPT-2 models on these tasks. Experiments on two types of ODEs show that GPT-2 can effectively learn a \textit{meta-ODE algorithm}, with convergence behavior comparable to, or better than, the Euler method, and achieve exponential accuracy gains with increasing numbers of demonstrations. Moreover, the model generalizes to out-of-distribution (OOD) problems, demonstrating robust extrapolation capabilities. These empirical findings provide new insights into the mechanisms of ICL in NLP and its potential for solving nonlinear numerical problems.
\end{abstract}

\section{Introduction}

\begin{figure*}[t]
    \centering
    \includegraphics[width=0.259\linewidth]{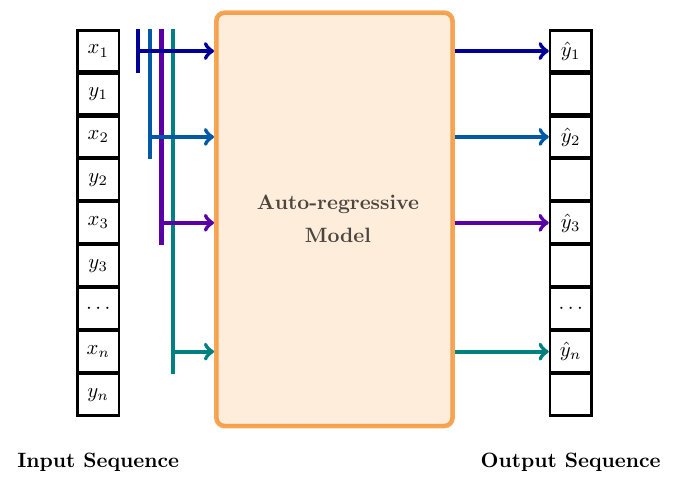}\hfill
    \includegraphics[width=0.247\linewidth]{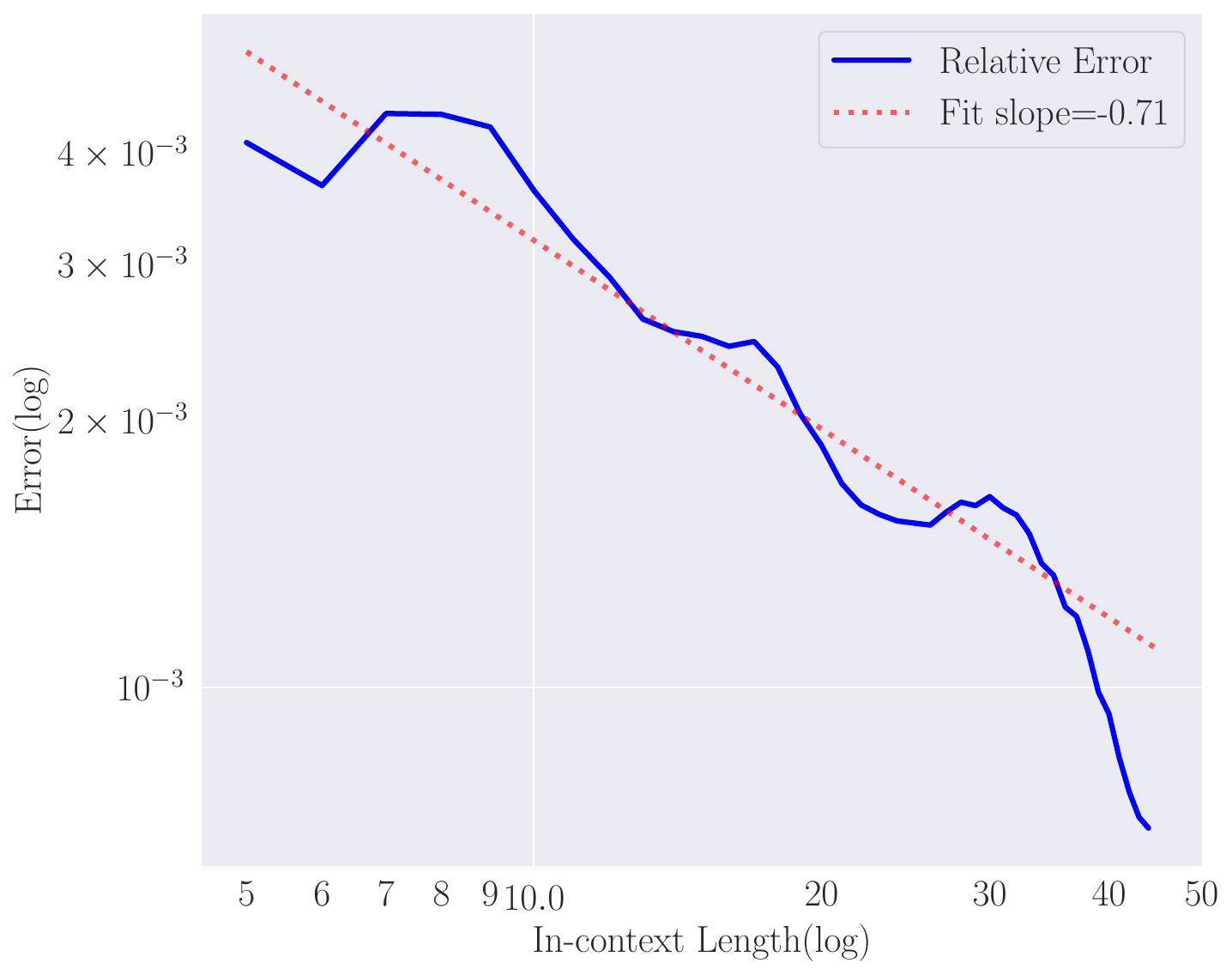}\hfill
    \includegraphics[width=0.247\linewidth]{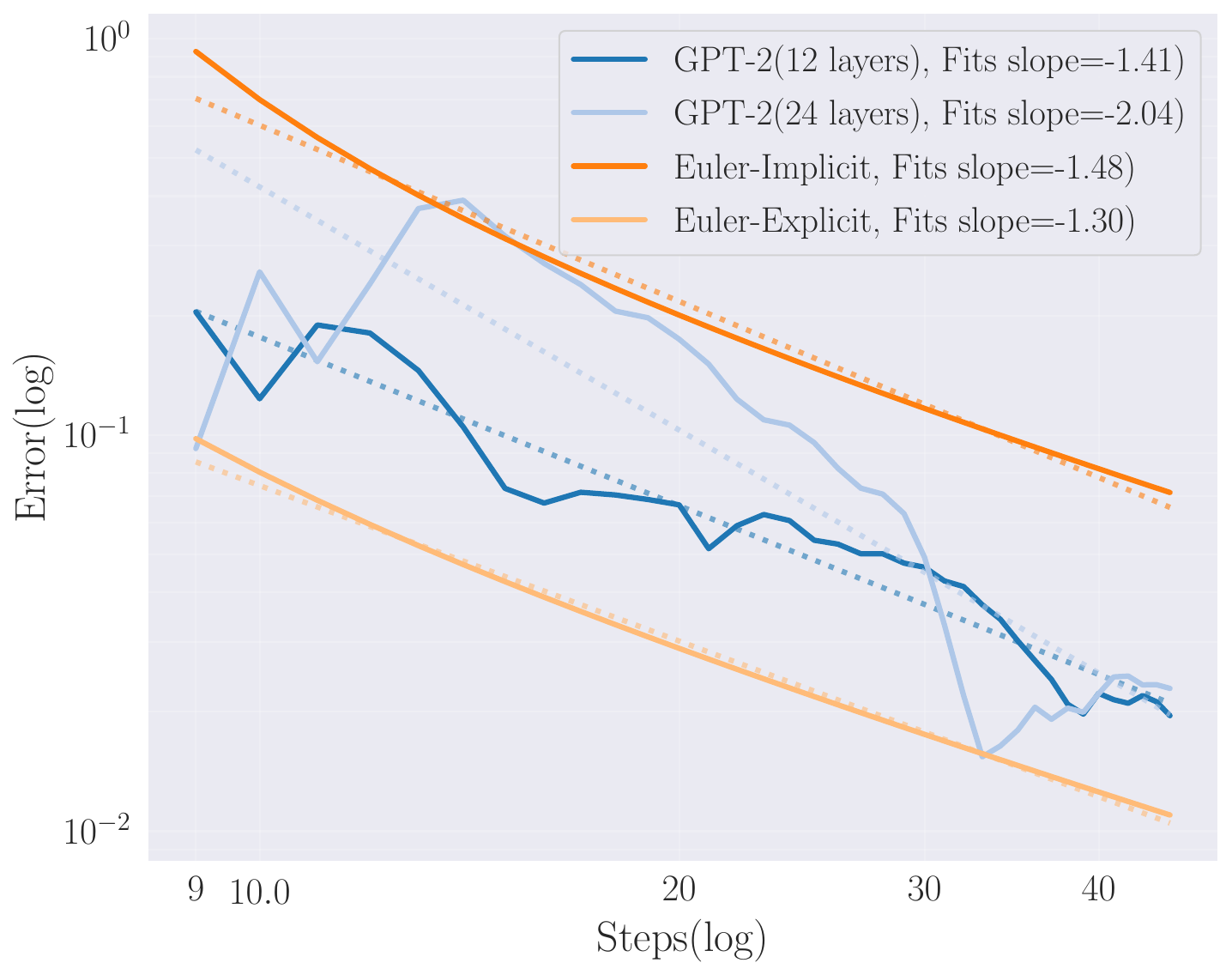}\hfill
    \includegraphics[width=0.247\linewidth]{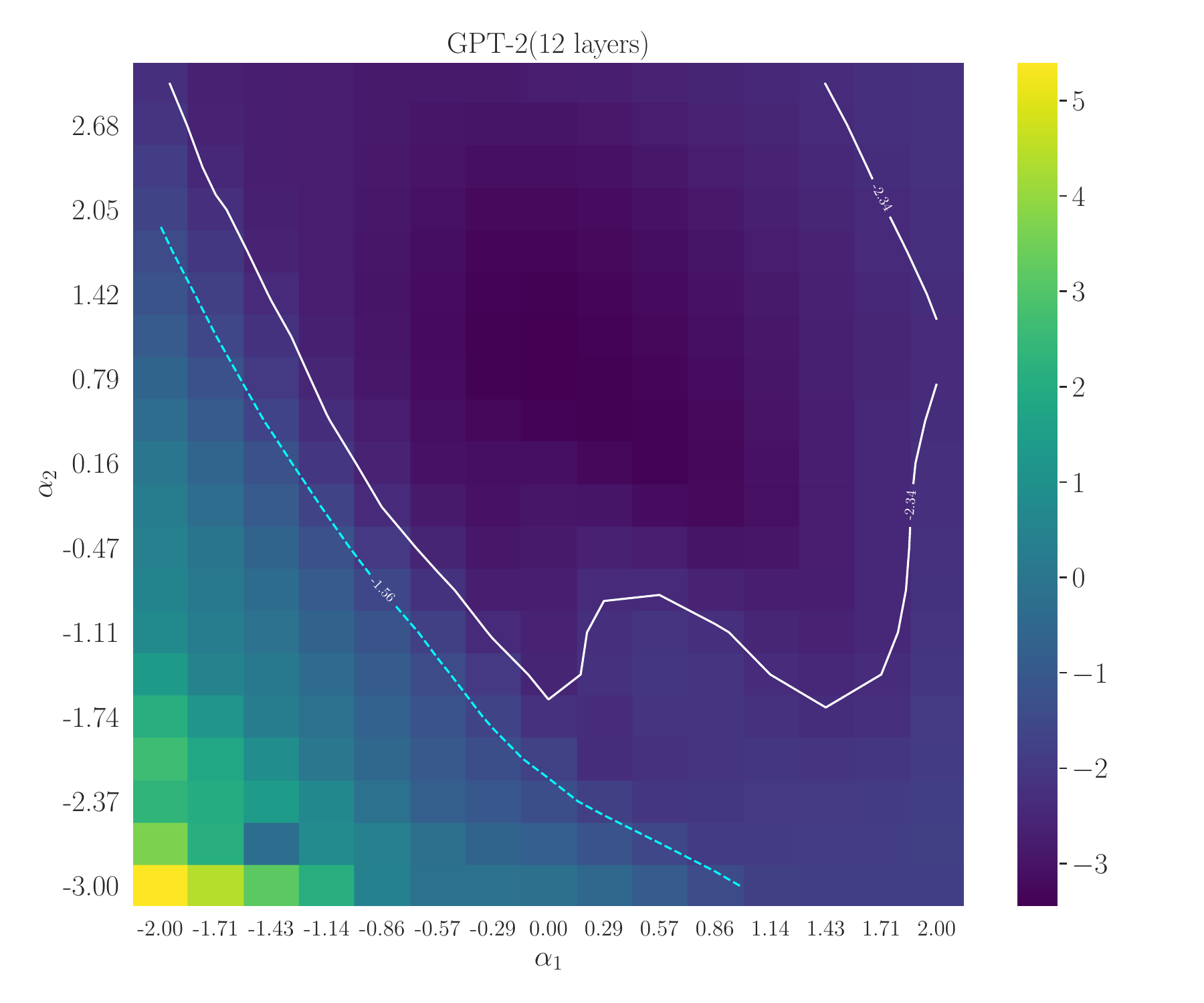}
    \caption{
ICL architecture and experimental results.\textit{Outer Left}: Illustration of ICL setup \citep{garg2022can}; \textit{Center left}: Error variation curve as a function of context length; \textit{Center right}: Comparison of GPT-2 predictions with classical methods; \textit{Outer Right}: Generalization error region plotted for GPT-2 (12 layers), where $\alpha_1$ and $\alpha_2$ are ODE parameters scaled from the training distribution (1.5$\times$). The dark region indicates a certain level of generalization. \vspace{-5mm}}
\label{fig:01-main-findings}
\end{figure*}

\textit{In-context learning} (ICL) \cite{brown2020language} has emerged as a pivotal feature among the capabilities of LLMs. It enables models to learn effectively through contextual prompts composed of input-output pairs without relying on parameter updates \cite{Anil2023PaLM2T,10434081}. This in-context learning ability is credited to \textit{emergent abilities} \cite{wei2022emergent,lu-etal-2024-emergent} of these Transformer-based LLMs \cite{Vaswani2017AttentionIA}. However, it is still unclear why or what these models can learn new tasks with only a few pairs of demonstrations. 

Recent studies \cite{garg2022can,xie2022an,von2023transformers,vladymyrov2024linear,fu2024transformers} have explored the mechanisms of ICL, primarily focusing on linear regression tasks. These works demonstrate that trained Transformer models can achieve efficiency comparable to classic methods under the ICL setting. In particular, models trained on linear examples have been shown to mimic gradient descent \cite{von2023transformers} and even higher-order optimization methods \cite{vladymyrov2024linear,fu2024transformers}. However, these findings primarily focus on linear patterns or simplified problems, leaving the behavior of full nonlinear Transformer models, especially in inherently nonlinear settings like NLP, insufficiently understood.

This work investigates the applicability of ICL to nonlinear numerical problems, extending its scope to the domain of ordinary differential equations (ODEs) and examining its potential in inherently nonlinear settings. 
We show that \textit{language models (e.g., GPT-2 models) can effectively learn meta-ODE solvers and exhibit strong generalization to new ODEs.} Our main contributions are summarized as follows\footnote{Our codes are available at \url{https://github.com/Zephyr-ziyang/GPT2_as_ODE_Solver}}:

\begin{itemize}
\item We design a tailored ICL framework for solving ODEs by encoding these nonlinear ODEs into parameterized sequence prompts. This formulation enables Transformers (e.g., GPT-2) to learn the underlying dynamics, achieving performance comparable to explicit and implicit Euler methods and surpassing them in some cases (see Figure \ref{fig:01-main-findings} center right). 

\item We then demonstrate out-of-distribution (OOD) generalization of ICL in ODE solving. The framework exhibits robustness to parameter distribution shifts, with deeper Transformer models showing stronger generalization capabilities (see Figure \ref{fig:01-main-findings} outer right).

\item We evaluate the stability of ICL-based solvers through multi-parameter extrapolation tests. Our results show that ICL achieves greater stability across wider parameter ranges compared to existing Euler methods. These findings suggest that Transformer-based models, while originally developed for NLP tasks, may also be capable of solving a broader class of numerical problems. Our preliminary results indicate that such models have the potential to serve as \textit{universal numerical solvers}.
\end{itemize}

\section{Methodology}

\noindent \textbf{Problem setup.} We study how GPT-2 models can learn to solve initial value problems (IVPs), where ordinary differential equations (ODEs) are defined by specified initial conditions (see Appendix~\ref{appendix:more-exp-details} for formal definitions). This framework enables a systematic analysis of temporal dynamics through parameterized differential equations. Specifically, we consider the following nonlinear task: each training prompt encodes a task comprising $N$ example pairs ${(\vx_i, \vy_i)}_{i=1}^N$, where $\vx_i \in \R^d$ stores the ODE parameters and $\vy_i \in \R^d$ represents the corresponding $n$-step solution. Each input $\vx_i = (\text{Para}_i, t_e, \text{Steps}_i)$ encodes the equation’s parameters, final time $t_e$, and number of time steps. The output $\vy_i = (\vy_i(t_j))$ contains the ODE solution sampled at discrete time points $t_j = 0, \dots, t_e$ for $j = 1, \dots, \text{Steps}_i$. We apply zero-padding to standardize outputs.

\noindent\textbf{Training loss.} At inference time, the model exhibits in-context learning (ICL) when it predicts $\hat{\vy}_q \approx h(\vx_q)$ without any weight updates, by leveraging the contextual examples in the prompt. %To encourage this behavior during training, we use a sliced mean squared error (sliced-MSE) loss:
We performed zero-padding on sample data exceeding the required number of steps. To prevent the model from learning the padded zeros, we use a sliced mean squared error (sliced-MSE) loss:
\begin{equation}
\label{equation: sliced_mse}
{\ell}(\vy, \hat{\vy}) = \frac{1}{N} \sum_{i=1}^N \left\|\vy(:i) - \hat{\vy}(:i)\right\|_2^2,
\end{equation}
where $\vy(:i)$ and $\hat{\vy}(:i)$ denote the first $i$ entries of $\vy$ and $\hat{\vy}$, respectively.

\noindent\textbf{Experimental setups.} We primarily follow the experimental setups from \citet{garg2022can}. For each experiment, we apply curriculum learning \cite{wang2021survey}, gradually expanding context length to 41 and vector dimensionality to 64 over the first 30k training steps. All models were trained for 600k steps before evaluation. We use AdamW \cite{kingmaB14} and employ an adjusted cosine annealing schedule for optimization. 

\section{Experiment Results}
\label{chap:experiment}

% Leveraging Transformers' success in linear regression \cite{von2023transformers}, we extend their capabilities to nonlinear differential equation solving via prompt sequence design. Focusing on initial value problems (IVPs) as a foundational case, we develop tailored prompt structures for differential equations, conducting comparative analysis with classical solvers to assess in-context learning efficacy.

% \subsection{Transformer Achieves Comparable Results to Euler Methods}
\subsection{ICL Matches Euler Methods}
\label{sec: further trial}

Building on the promising results of the GPT-2 model in solving basic differential equations (an initial trial is provided in Appendix~\ref{appendix: first trail}, where a 24-layer GPT-2 is introduced alongside our original 12-layer model), we extend our investigation to initial value problems (IVPs), specifically first-order linear ODEs with five degrees of parametric freedom (see Appendix~\ref{def:1_order_ivp} for the formal definition). To evaluate model performance more concretely, we conduct comparative experiments with classical numerical solvers, including both the Euler-Explicit and Euler-Implicit methods.

\begin{figure}[htpb]
    \includegraphics[width=0.49\linewidth]{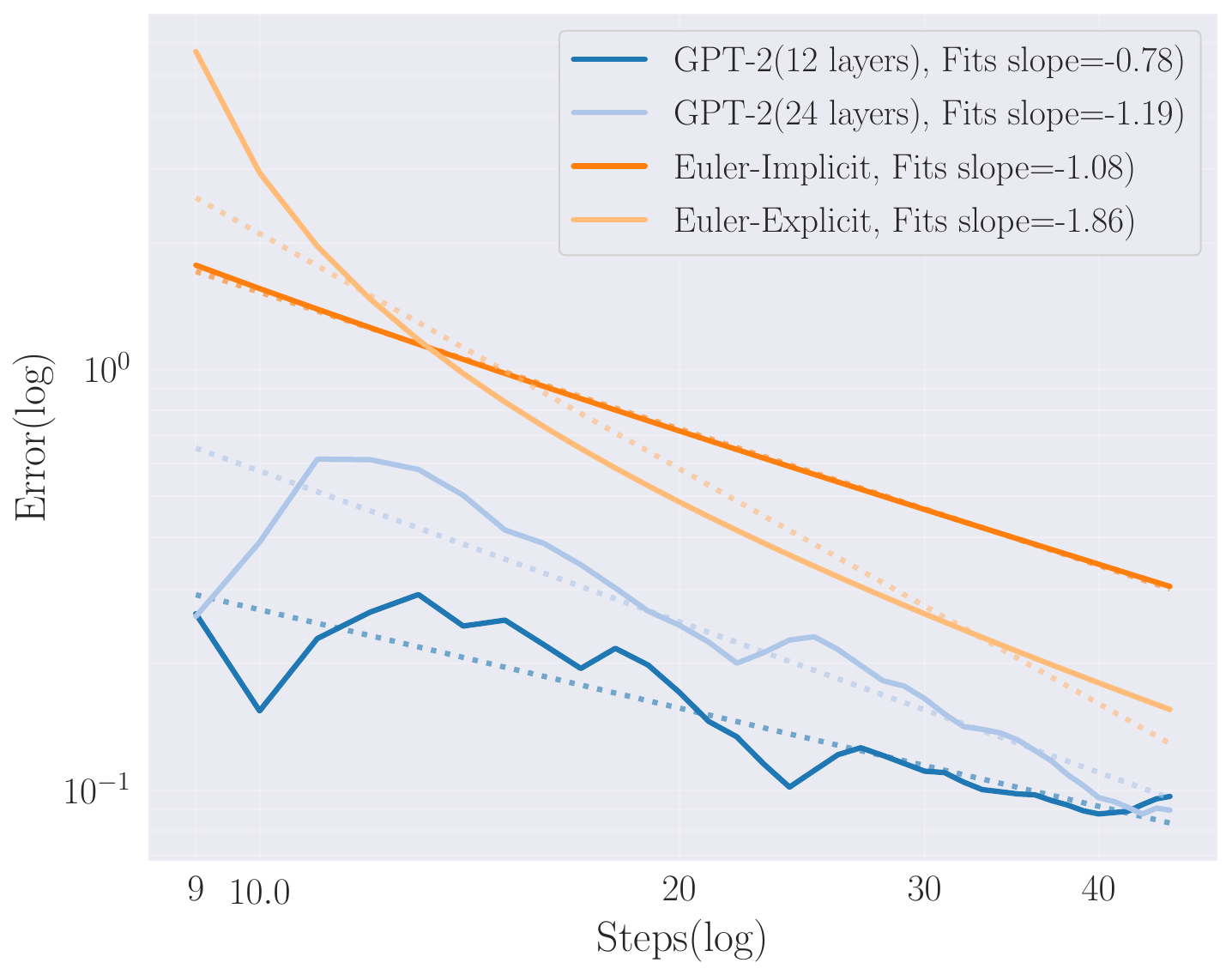}
    \includegraphics[width=0.49\linewidth]{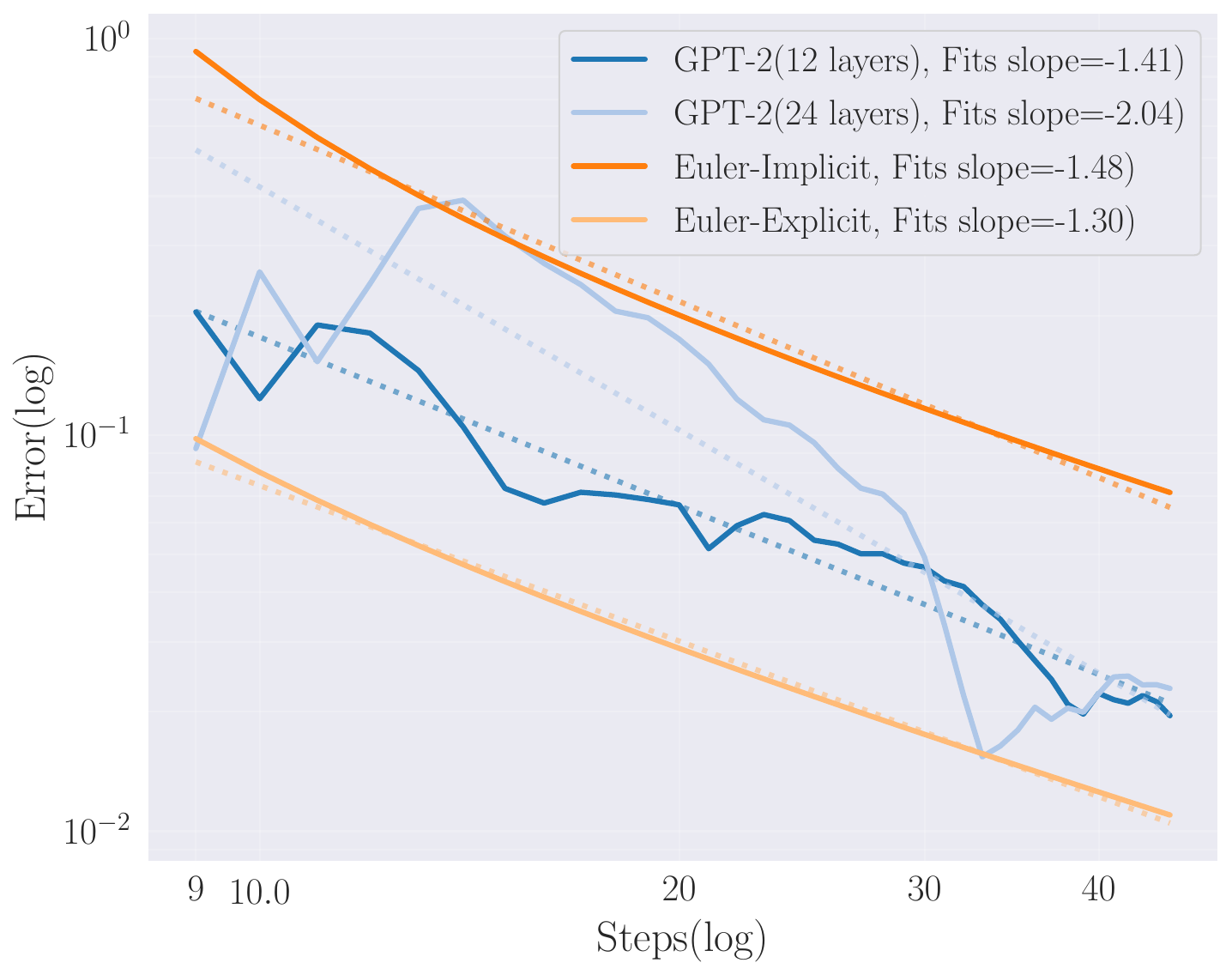}
    \caption{Log-log plots comparing GPT-2 and Euler methods. \textit{Left}: GPT-2 outperforms Euler methods. \textit{Right}: Comparable performance between GPT-2 and Euler.
\textit{Steps} denote context length for GPT-2 and iteration steps for Euler methods.}
    \label{fig:with_classic}
\end{figure}

\begin{figure*}[tpb!]
    \centering
    \includegraphics[width=0.9\linewidth,height=0.2\linewidth]{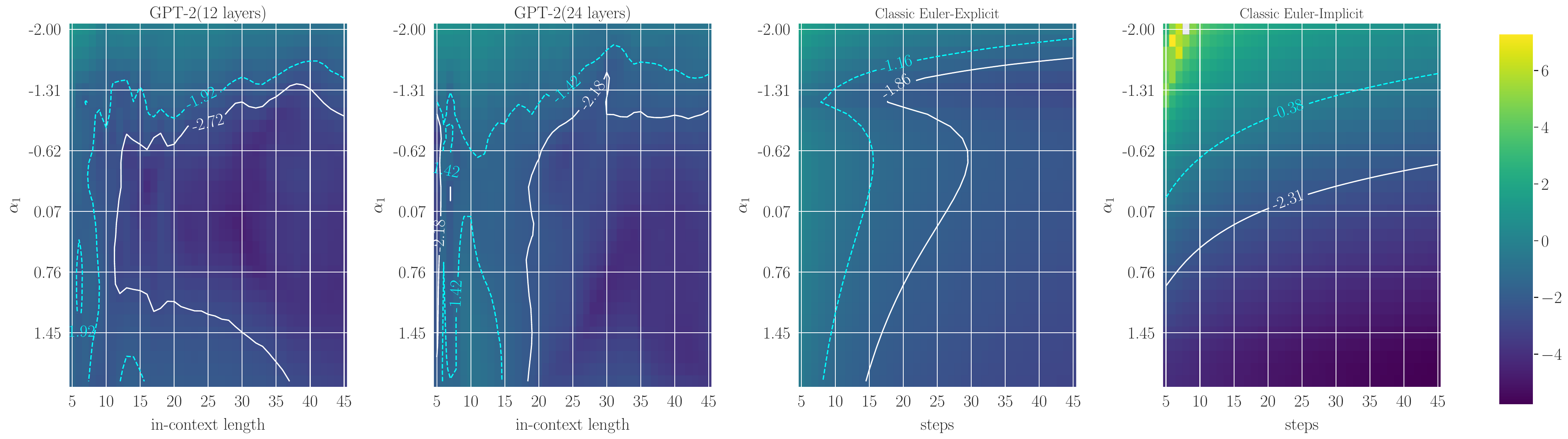}
    \caption{
        Heatmap of Solution error comparison under $\alpha_1$ shifts (extending original training range: $\alpha_1\in [-1, 1]$ towards $[-2,2]$). \textit{From left}: 12L/24L GPT-2, explicit/implicit Euler. 
        % (left): 12-layer GPT-2; (center left): 24-layer GPT-2; 
        % (center right): Explicit Euler; (right): Implicit Euler. 
        For traditional methods, context length aligns with iteration steps for comparison. Contour lines mark 50\% (cyan) and 70\% (white) of each subplot's error range.
     }
    \label{fig:Robustness of a_1}
\end{figure*}

\begin{figure*}[tpb!]
    \centering
    \includegraphics[width=0.9\linewidth, height=0.2\linewidth]{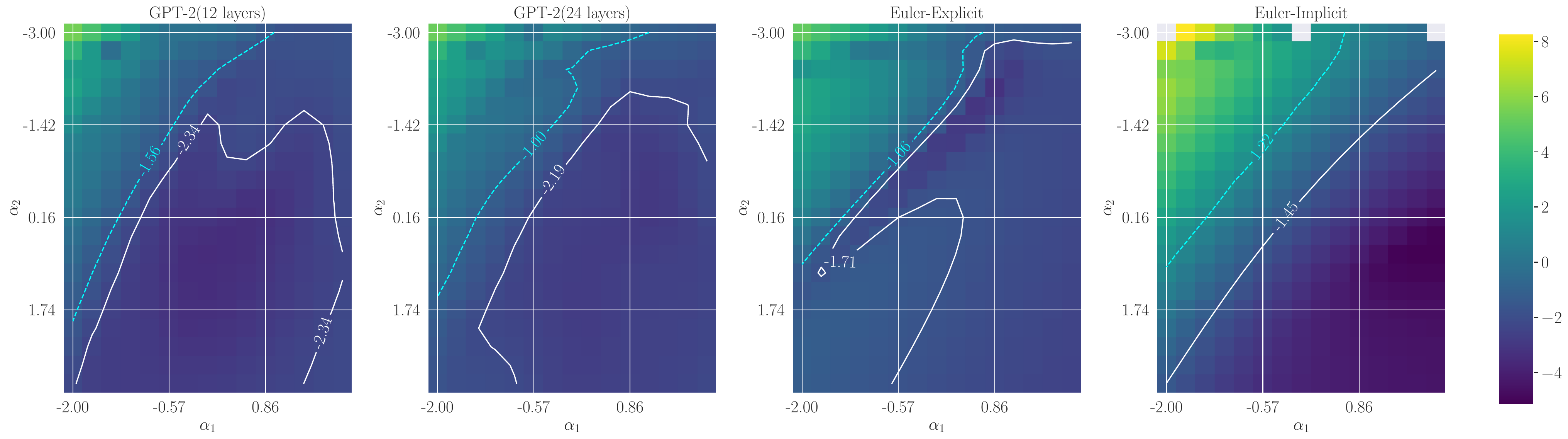}
\caption{
Heatmap of error region comparisons across parameter combinations.  $\alpha_1$-$\alpha_2$ (trained on $[-1, 1]\times[-2, 2]$, tested on $[-2, 2]\times[-3, 3]$). \textit{From left}: 12L/24L GPT-2, explicit/implicit Euler. Contours mark 50\% (cyan) and 70\% (white) of each subplot's range.}
\label{fig:a_1a_2 error  heatmap}
\end{figure*}

The model’s performance is shown in Figure~\ref{fig:with_classic}. Our benchmarks against the Euler methods reveal that the Transformer outperforms classical approaches in some cases while achieving comparable accuracy in others.

GPT-2 models demonstrate dual advantages in both accuracy and adaptability compared to explicit Euler methods. Specifically, well-trained models achieve comparable or superior accuracy to Euler integration while maintaining better numerical stability, and additionally exhibit stronger adaptability when handling stiff differential equations or conducting long-term integration.

Notably, deeper architectures (24-layer vs. 12-layer) demonstrate diminishing returns when scaling depth, attaining only marginal accuracy gains despite doubled parameters – a saturation pattern consistent with findings in \citet{fu2024transformers}. As expected, GPT-2 models struggle to match classical solvers in scenarios requiring high-precision solutions, reflecting fundamental limitations of neural approximators rather than implementation flaws (see more discussions in the Appendix~\ref{appendix: limitations on zero case}).

\subsection{ICL Generalizes across Distributions}

Building on the previous subsection, which demonstrates that GPT-2 models can effectively solve first-order ODEs as shown in Figure~\ref{fig:with_classic}, we further evaluate the generalization capabilities of GPT-2 by extending parameter ranges beyond the training distribution as illustrated in Figure~\ref{fig:Robustness of a_1}.

Our experimental results indicate that the model maintains stability while parameter range shifting, representing generalization ability. Within the original $\alpha_1 \in [-1,1]$ range, errors monotonically decrease as context length increases. Beyond this distribution ($\alpha_1 \in [-2,2]$), it preserves reasonable accuracy that still improves with longer contexts. Compared to Euler's stepwise error decay, GPT-2 models exhibit smooth error convergence through adaptive in-context learning.

Notably, the 24-layer variant shows better extrapolation on negative $\alpha_1$ despite overall lower accuracy than the 12-layer model, suggesting depth impacts generalization patterns.

\begin{figure*}[htpb]
    \centering
    \includegraphics[width=0.9\linewidth, height=0.2\linewidth]{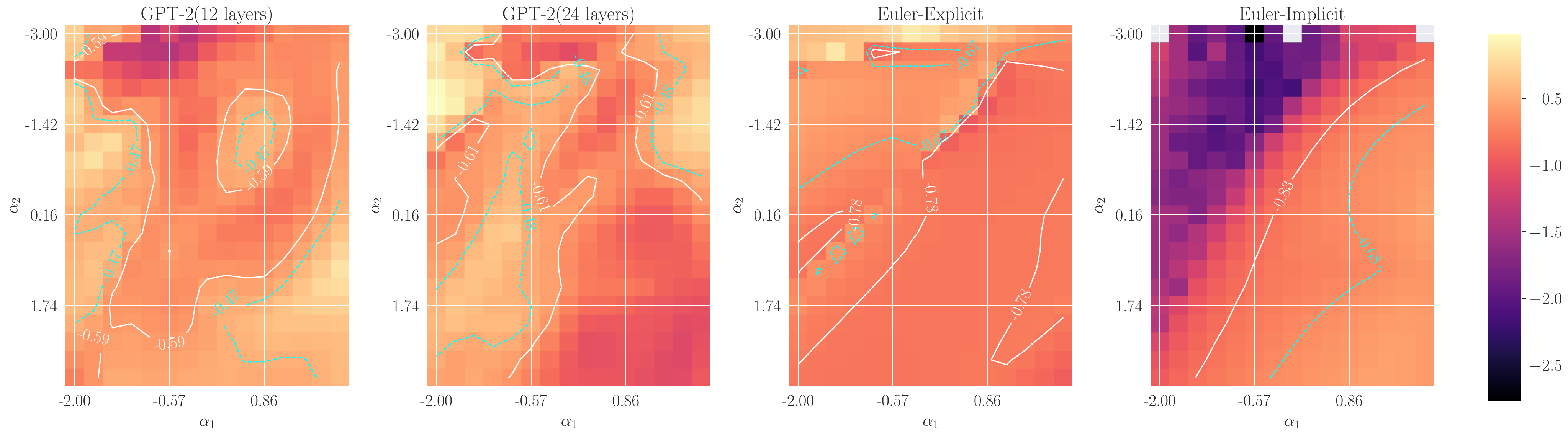}
\caption{Heatmap of convergence slope comparisons. Models were trained on parameter ranges $\alpha_1 \times \alpha_2 \in [-1, 1] \times [-2, 2]$ and tested on a broader range of $\alpha_1 \times \alpha_2 \in [-2, 2] \times [-3, 3]$. Contour lines indicate 50\% (cyan) and 70\% (white) levels relative to the maximum value in each subplot.}
    \label{fig:a_1a_2 slope  heatmap}
\end{figure*}

\subsection{ICL is Relative Stable to Classics}

As Figure~\ref{fig:with_classic} reveals precision degradation in Euler methods when handling stiff problems (attributed to step size adaptation limitations), we conduct systematic cross-testing across dual parameter axes to assess whether GPT-2 with ICL exhibits analogous instability patterns.

\noindent \textbf{Error stability regions.}  
Under fixed context length and iteration steps (45 precisely), test results are shown in Figure~\ref{fig:a_1a_2 error  heatmap}. The error distributions with stable error regions exhibited by GPT-2 model are comparable to or larger than Euler methods. This likely stems from its adaptive capability—adjusting learning strategies based on context length and iteration steps to maintain high precision. In contrast, explicit Euler's fixed-step mechanism becomes suboptimal with parameter variations, leading to accuracy deterioration.

Beyond the $\alpha$ parameter performance, the Transformer also demonstrates enhanced adaptation capability under $\beta_1$-$\beta_2$ shifts compared to Euler methods (Appendix~\ref{appendix: more composite}), while exhibiting precision variations. This phenomenon can be explained by the model's inherent sensitivity to input ordering, where consistently configured $\alpha$ parameters predominantly influence the attention mechanisms \cite{zhao2021calibrate, lu2022fantastically, Chan2022TransformersGD}.

\noindent \textbf{Convergence slope stability.}  When context length and iteration steps grow, we reveal a exponential convergence. The dynamic of convergence slope is shown in Figure~\ref{fig:a_1a_2 slope  heatmap}. Compared to error maps, slope maps exhibit greater instability: Euler methods maintain steady convergence rates while GPT-2 show localized volatility with sudden low-slope valleys.

We attribute this performance to: (1) Slope, as a global estimator of error reduction, amplifies instability from probabilistic solutions, whereas Euler methods only show slope degradation when step sizes are insufficient (implicit Euler demonstrates better stability); (2) For extreme parameters, in-context learning fails to capture logical parameter relationships, weakening the error reduction trend with longer contexts.

Notably, deeper models (24L) display smoother slope transitions in $\alpha_1>1.5$ regions, suggesting depth may mitigate convergence instability in specific parameter ranges. However, this improvement is selective—deeper models show exacerbated volatility in $\alpha_2$'s negative half-axis.

\subsection{Comparative Performance Analysis} \label{sec:comparative}
As shown in Table~\ref{tab:comparison-convergence}, the core advantages of GPT-2 models lie in adaptive convergence and global modeling capabilities. Compared to the $\mathcal{O}(h)$ convergence of explicit Euler and $\mathcal{O}(h^2)$ of implicit Euler, 12-layer and 24-layer GPT-2 models achieved exponential convergence rates of $\mathcal{O}(e^{-kN})$ and $\mathcal{O}(e^{-k'N})$ respectively. We hypothesize exponential convergence of model accuracy with ICL length (See Appendix \ref{appendix: conj}, Conjecture \ref{conj: exponential}). This difference stems from the attention mechanism's comprehensive utilization of historical information, enabling implicit variable-step strategies. Notably, when handling stiff equations, traditional methods often require frequent parameter adjustments due to fixed-step limitations, whereas GPT-2 exhibited smoother error reduction curves.

\begin{table}[htbp]
\small
  \centering
%\caption{Performance Comparison Between Transformers and Euler Methods}
  \begin{tabular}{lccc}
    \toprule
    % \hline
    Method & Convergence Rate & Generalization\\
    \midrule
    % \hline
    Euler-Explicit & $\mathcal{O}(h)$ & Low\\
    Euler-Implicit & $\mathcal{O}(h^2)$ & Medium  \\
    GPT-2 ($L=12$) & $\mathcal{O}(e^{-kN})$ & Medium-High\\
    GPT-2 ($L=24$) & $\mathcal{O}(e^{-k'N})$ & High\\
    \bottomrule
    % \hline
  \end{tabular}
  \caption{Comparison between GPT-2 models and traditional explicit/implicit Euler methods across various metrics.}
  \label{tab:comparison-convergence}
\end{table}

\section{Conclusions and discussions}

% \subsection{Analysis and Prospects of In-Context Learning Model Performance}

\label{chap:conclusion}

This study demonstrates that GPT-2 models can effectively solve ordinary differential equations through in-context learning, offering three key findings: (1) Exponential error decay with increasing context length via adaptive convergence mechanisms, (2) Generalization capability under extended parameter distributions, with maintained convergence efficacy as context length grows, and (3) Preserved convergence rates during parameter extrapolation, despite increased volatility in slope stability.   Experiments confirm that the model maintains numerical stability comparable to Euler methods for nonlinear numerical problem: ODE numerical solution, while exhibiting promising generalization under parameter distribution shifts.

\clearpage

\section*{Limitations}

Our results may have the following limitations: current observations are from only GPT-2 models. Larger model configurations could be conducted.

We conducted a conjecture, waiting for theoretical analysis of Transformers' internal mechanisms for learning differential equations, internal impact from optimization of in-context learning strategies (e.g., positional encoding and attention masking). While the model’s parallel prediction offers significant advantages for real-time simulations, .

\textbf{Ethical considerations.} As our research involves synthetic mathematical data and does not engage with human subjects, sensitive content, or real-world applications, we do not foresee direct ethical risks.

\bibliography{custom}

\begin{thebibliography}{32}
\providecommand{\natexlab}[1]{#1}

\bibitem[{Anil et~al.(2023)Anil, Dai, Firat, Johnson, Lepikhin, Passos, Shakeri, Taropa, Bailey, Chen, Chu, Clark, Shafey, Huang, Meier-Hellstern, Mishra, Moreira, Omernick, Robinson, Ruder, Tay, Xiao, Xu, Zhang, Abrego, Ahn, Austin, Barham, Botha, Bradbury, Brahma, Brooks, Catasta, Cheng, Cherry, Choquette-Choo, Chowdhery, Crepy, Dave, Dehghani, Dev, Devlin, Díaz, Du, Dyer, Feinberg, Feng, Fienber, Freitag, Garcia, Gehrmann, Gonzalez, Gur-Ari, Hand, Hashemi, Hou, Howland, Hu, Hui, Hurwitz, Isard, Ittycheriah, Jagielski, Jia, Kenealy, Krikun, Kudugunta, Lan, Lee, Lee, Li, Li, Li, Li, Li, Lim, Lin, Liu, Liu, Maggioni, Mahendru, Maynez, Misra, Moussalem, Nado, Nham, Ni, Nystrom, Parrish, Pellat, Polacek, Polozov, Pope, Qiao, Reif, Richter, Riley, Ros, Roy, Saeta, Samuel, Shelby, Slone, Smilkov, So, Sohn, Tokumine, Valter, Vasudevan, Vodrahalli, Wang, Wang, Wang, Wang, Wieting, Wu, Xu, Xu, Xue, Yin, Yu, Zhang, Zheng, Zheng, Zhou, Zhou, Petrov, and Wu}]{Anil2023PaLM2T}
Rohan Anil, Andrew~M. Dai, Orhan Firat, Melvin Johnson, Dmitry Lepikhin, Alexandre Passos, Siamak Shakeri, Emanuel Taropa, Paige Bailey, Zhifeng Chen, Eric Chu, Jonathan~H. Clark, Laurent~El Shafey, Yanping Huang, Kathy Meier-Hellstern, Gaurav Mishra, Erica Moreira, Mark Omernick, Kevin Robinson, and 109 others. 2023.
\newblock \href {https://arxiv.org/abs/2305.10403} {Palm 2 technical report}.
\newblock \emph{Preprint}, arXiv:2305.10403.

\bibitem[{Bai et~al.(2023)Bai, Chen, Wang, Xiong, and Mei}]{NEURIPS2023_b2e63e36}
Yu~Bai, Fan Chen, Huan Wang, Caiming Xiong, and Song Mei. 2023.
\newblock \href {https://proceedings.neurips.cc/paper_files/paper/2023/hash/b2e63e36c57e153b9015fece2352a9f9-Abstract-Conference.html} {Transformers as statisticians: Provable in-context learning with in-context algorithm selection}.
\newblock \emph{Advances in Neural Information Processing Systems}, 36:57125--57211.

\bibitem[{Bengio et~al.(2009)Bengio, Louradour, Collobert, and Weston}]{bengio2009curriculum}
Yoshua Bengio, J{\'e}r{\^o}me Louradour, Ronan Collobert, and Jason Weston. 2009.
\newblock \href {https://dl.acm.org/doi/abs/10.1145/1553374.1553380} {Curriculum learning}.
\newblock In \emph{Proceedings of the 26th annual international conference on machine learning}, pages 41--48.

\bibitem[{Bondaschi et~al.(2025)Bondaschi, Rajaraman, Wei, Ramchandran, Pascanu, Gulcehre, Gastpar, and Makkuva}]{bondaschi2025from}
Marco Bondaschi, Nived Rajaraman, Xiuying Wei, Kannan Ramchandran, Razvan Pascanu, Caglar Gulcehre, Michael Gastpar, and Ashok~Vardhan Makkuva. 2025.
\newblock \href {https://openreview.net/forum?id=Qwt5andjuU} {From markov to laplace: How mamba in-context learns markov chains}.
\newblock In \emph{ICLR 2025 Workshop: XAI4Science: From Understanding Model Behavior to Discovering New Scientific Knowledge}.

\bibitem[{Brown et~al.(2020)Brown, Mann, Ryder, Subbiah, Kaplan, Dhariwal, Neelakantan, Shyam, Sastry, Askell et~al.}]{brown2020language}
Tom~B Brown, Benjamin Mann, Nick Ryder, Melanie Subbiah, Jared Kaplan, Prafulla Dhariwal, Arvind Neelakantan, Pranav Shyam, Girish Sastry, Amanda Askell, and 1 others. 2020.
\newblock \href {https://proceedings.neurips.cc/paper/2020/hash/1457c0d6bfcb4967418bfb8ac142f64a-Abstract.html} {Language models are few-shot learners}.
\newblock In \emph{Proceedings of the 34th International Conference on Neural Information Processing Systems}, pages 1877--1901.

\bibitem[{Chan et~al.(2022)Chan, Dasgupta, Kim, Kumaran, Lampinen, and Hill}]{Chan2022TransformersGD}
Stephanie C.~Y. Chan, Ishita Dasgupta, Junkyung Kim, Dharshan Kumaran, Andrew~K. Lampinen, and Felix Hill. 2022.
\newblock \href {https://arxiv.org/abs/2210.05675} {Transformers generalize differently from information stored in context vs in weights}.
\newblock \emph{Preprint}, arXiv:2210.05675.

\bibitem[{Chen et~al.(2024)Chen, Sheen, Wang, and Yang}]{chen2024unveiling}
Siyu Chen, Heejune Sheen, Tianhao Wang, and Zhuoran Yang. 2024.
\newblock \href {https://openreview.net/forum?id=4fN2REs0Ma} {Unveiling induction heads: Provable training dynamics and feature learning in transformers}.
\newblock In \emph{The Thirty-eighth Annual Conference on Neural Information Processing Systems}.

\bibitem[{DeepSeek-AI et~al.(2025)DeepSeek-AI, Liu, Feng, Xue, Wang, Wu, Lu, Zhao, Deng, Zhang, Ruan, Dai, Guo, Yang, Chen, Ji, Li, Lin, Dai, Luo, Hao, Chen, Li, Zhang, Bao, Xu, Wang, Zhang, Ding, Xin, Gao, Li, Qu, Cai, Liang, Guo, Ni, Li, Wang, Chen, Chen, Yuan, Qiu, Li, Song, Dong, Hu, Gao, Guan, Huang, Yu, Wang, Zhang, Xu, Xia, Zhao, Wang, Zhang, Li, Wang, Zhang, Zhang, Tang, Li, Tian, Huang, Wang, Zhang, Wang, Zhu, Chen, Du, Chen, Jin, Ge, Zhang, Pan, Wang, Xu, Zhang, Chen, Li, Lu, Zhou, Chen, Wu, Ye, Ye, Ma, Wang, Zhou, Yu, Zhou, Pan, Wang, Yun, Pei, Sun, Xiao, Zeng, Zhao, An, Liu, Liang, Gao, Yu, Zhang, Li, Jin, Wang, Bi, Liu, Wang, Shen, Chen, Zhang, Chen, Nie, Sun, Wang, Cheng, Liu, Xie, Liu, Yu, Song, Shan, Zhou, Yang, Li, Su, Lin, Li, Wang, Wei, Zhu, Zhang, Xu, Xu, Huang, Li, Zhao, Sun, Li, Wang, Yu, Zheng, Zhang, Shi, Xiong, He, Tang, Piao, Wang, Tan, Ma, Liu, Guo, Wu, Ou, Zhu, Wang, Gong, Zou, He, Zha, Xiong, Ma, Yan, Luo, You, Liu, Zhou, Wu, Ren, Ren, Sha, Fu, Xu, Huang, Zhang, Xie, Zhang, Hao,
  Gou, Ma, Yan, Shao, Xu, Wu, Zhang, Li, Gu, Zhu, Liu, Li, Xie, Song, Gao, and Pan}]{DBLP:journals/corr/abs-2412-19437}
DeepSeek-AI, Aixin Liu, Bei Feng, Bing Xue, Bingxuan Wang, Bochao Wu, Chengda Lu, Chenggang Zhao, Chengqi Deng, Chenyu Zhang, Chong Ruan, Damai Dai, Daya Guo, Dejian Yang, Deli Chen, Dongjie Ji, Erhang Li, Fangyun Lin, Fucong Dai, and 181 others. 2025.
\newblock \href {https://arxiv.org/abs/2412.19437} {Deepseek-v3 technical report}.
\newblock \emph{Preprint}, arXiv:2412.19437.

\bibitem[{Dong et~al.(2024)Dong, Li, Dai, Zheng, Ma, Li, Xia, Xu, Wu, Chang et~al.}]{dong2024survey}
Qingxiu Dong, Lei Li, Damai Dai, Ce~Zheng, Jingyuan Ma, Rui Li, Heming Xia, Jingjing Xu, Zhiyong Wu, Baobao Chang, and 1 others. 2024.
\newblock \href {https://aclanthology.org/2024.emnlp-main.64/} {A survey on in-context learning}.
\newblock In \emph{Proceedings of the 2024 Conference on Empirical Methods in Natural Language Processing}, pages 1107--1128.

\bibitem[{Edelman et~al.(2024)Edelman, Tsilivis, Edelman, Malach, and Goel}]{edelman2024evolution}
Ezra Edelman, Nikolaos Tsilivis, Benjamin Edelman, Eran Malach, and Surbhi Goel. 2024.
\newblock \href {https://proceedings.neurips.cc/paper_files/paper/2024/hash/75b0edb869e2cd509d64d0e8ff446bc1-Abstract-Conference.html} {The evolution of statistical induction heads: In-context learning markov chains}.
\newblock \emph{Advances in Neural Information Processing Systems}, 37:64273--64311.

\bibitem[{Elhage et~al.(2021)Elhage, Nanda, Olsson, Henighan, Joseph, Mann, Askell, Bai, Chen, Conerly, DasSarma, Drain, Ganguli, Hatfield-Dodds, Hernandez, Jones, Kernion, Lovitt, Ndousse, Amodei, Brown, Clark, Kaplan, McCandlish, and Olah}]{elhage2021mathematical}
Nelson Elhage, Neel Nanda, Catherine Olsson, Tom Henighan, Nicholas Joseph, Ben Mann, Amanda Askell, Yuntao Bai, Anna Chen, Tom Conerly, Nova DasSarma, Dawn Drain, Deep Ganguli, Zac Hatfield-Dodds, Danny Hernandez, Andy Jones, Jackson Kernion, Liane Lovitt, Kamal Ndousse, and 6 others. 2021.
\newblock \href {https://transformer-circuits.pub/2021/framework/index.html} {A mathematical framework for transformer circuits}.
\newblock \emph{Transformer Circuits Thread}.

\bibitem[{Fu et~al.(2024)Fu, Chen, Jia, and Sharan}]{fu2024transformers}
Deqing Fu, Tian-qi Chen, Robin Jia, and Vatsal Sharan. 2024.
\newblock \href {https://proceedings.neurips.cc/paper_files/paper/2024/hash/b2d4051f03a7038a2771dfbbe5c7b54e-Abstract-Conference.html} {Transformers learn to achieve second-order convergence rates for in-context linear regression}.
\newblock \emph{Advances in Neural Information Processing Systems}, 37:98675--98716.

\bibitem[{Garg et~al.(2022)Garg, Tsipras, Liang, and Valiant}]{garg2022can}
Shivam Garg, Dimitris Tsipras, Percy~S Liang, and Gregory Valiant. 2022.
\newblock \href {https://proceedings.neurips.cc/paper_files/paper/2022/hash/c529dba08a146ea8d6cf715ae8930cbe-Abstract-Conference.html} {What can transformers learn in-context? a case study of simple function classes}.
\newblock \emph{Advances in Neural Information Processing Systems}, 35:30583--30598.

\bibitem[{Grazzi et~al.(2024)Grazzi, Siems, Schrodi, Brox, and Hutter}]{grazzi24mamba-icl}
Riccardo Grazzi, Julien~Niklas Siems, Simon Schrodi, Thomas Brox, and Frank Hutter. 2024.
\newblock \href {https://proceedings.mlr.press/v256/grazzi24a.html} {Is {Mamba} capable of in-context learning?}
\newblock In \emph{Proceedings of the Third International Conference on Automated Machine Learning}, volume 256 of \emph{Proceedings of Machine Learning Research}, pages 1/1--26. PMLR.

\bibitem[{Johnson et~al.(2023)Johnson, Xinying, Khaw, and Lee}]{johnson2023ps}
Olanrewaju~Victor Johnson, Chew Xinying, Khai~Wah Khaw, and Ming~Ha Lee. 2023.
\newblock \href {https://ieeexplore.ieee.org/abstract/document/10348553/} {ps-calr: periodic-shift cosine annealing learning rate for deep neural networks}.
\newblock \emph{IEEE Access}, 11:139171--139186.

\bibitem[{Kingma and Ba(2015)}]{kingmaB14}
Diederik~P. Kingma and Jimmy Ba. 2015.
\newblock \href {http://arxiv.org/abs/1412.6980} {Adam: A method for stochastic optimization}.
\newblock In \emph{ICLR (Poster)}.

\bibitem[{Liu et~al.(2025)Liu, Zhou, Shen, and Yang}]{liu2025learntooptimizecapabilitiestransformersincontext}
Renpu Liu, Ruida Zhou, Cong Shen, and Jing Yang. 2025.
\newblock \href {https://arxiv.org/abs/2410.13981} {On the learn-to-optimize capabilities of transformers in in-context sparse recovery}.
\newblock \emph{Preprint}, arXiv:2410.13981.

\bibitem[{Lu et~al.(2024)Lu, Bigoulaeva, Sachdeva, Tayyar~Madabushi, and Gurevych}]{lu-etal-2024-emergent}
Sheng Lu, Irina Bigoulaeva, Rachneet Sachdeva, Harish Tayyar~Madabushi, and Iryna Gurevych. 2024.
\newblock \href {https://aclanthology.org/2024.acl-long.279/} {Are emergent abilities in large language models just in-context learning?}
\newblock In \emph{Proceedings of the 62nd Annual Meeting of the Association for Computational Linguistics (Volume 1: Long Papers)}, pages 5098--5139.

\bibitem[{Lu et~al.(2022)Lu, Bartolo, Moore, Riedel, and Stenetorp}]{lu2022fantastically}
Yao Lu, Max Bartolo, Alastair Moore, Sebastian Riedel, and Pontus Stenetorp. 2022.
\newblock \href {https://aclanthology.org/2022.acl-long.556/} {Fantastically ordered prompts and where to find them: Overcoming few-shot prompt order sensitivity}.
\newblock In \emph{Proceedings of the 60th Annual Meeting of the Association for Computational Linguistics (Volume 1: Long Papers)}, pages 8086--8098.

\bibitem[{Olsson et~al.(2022)Olsson, Elhage, Nanda, Joseph, DasSarma, Henighan, Mann, Askell, Bai, Chen, Conerly, Drain, Ganguli, Hatfield-Dodds, Hernandez, Johnston, Jones, Kernion, Lovitt, Ndousse, Amodei, Brown, Clark, Kaplan, McCandlish, and Olah}]{olsson2022context}
Catherine Olsson, Nelson Elhage, Neel Nanda, Nicholas Joseph, Nova DasSarma, Tom Henighan, Ben Mann, Amanda Askell, Yuntao Bai, Anna Chen, Tom Conerly, Dawn Drain, Deep Ganguli, Zac Hatfield-Dodds, Danny Hernandez, Scott Johnston, Andy Jones, Jackson Kernion, Liane Lovitt, and 7 others. 2022.
\newblock \href {https://transformer-circuits.pub/2022/in-context-learning-and-induction-heads/index.html} {In-context learning and induction heads}.
\newblock \emph{Transformer Circuits Thread}.

\bibitem[{Ren et~al.(2024)Ren, Guo, Yan, Liu, Zhang, Qiu, and Lin}]{ren-etal-2024-identifying}
Jie Ren, Qipeng Guo, Hang Yan, Dongrui Liu, Quanshi Zhang, Xipeng Qiu, and Dahua Lin. 2024.
\newblock \href {https://aclanthology.org/2024.findings-acl.412/} {Identifying semantic induction heads to understand in-context learning}.
\newblock In \emph{Findings of the Association for Computational Linguistics: ACL 2024}, pages 6916--6932.

\bibitem[{Thakkar and Manimaran(2023)}]{10434081}
Hiren Thakkar and A~Manimaran. 2023.
\newblock \href {https://ieeexplore.ieee.org/abstract/document/10434081/} {Comprehensive examination of instruction-based language models: A comparative analysis of mistral-7b and llama-2-7b}.
\newblock In \emph{2023 International Conference on Emerging Research in Computational Science}, pages 1--6.

\bibitem[{Todd et~al.(2024)Todd, Li, Sharma, Mueller, Wallace, and Bau}]{todd2024functionvectorslargelanguage}
Eric Todd, Millicent~L. Li, Arnab~Sen Sharma, Aaron Mueller, Byron~C. Wallace, and David Bau. 2024.
\newblock \href {https://arxiv.org/abs/2310.15213} {Function vectors in large language models}.
\newblock \emph{Preprint}, arXiv:2310.15213.

\bibitem[{Varghese and Sambath(2024)}]{10533619}
Rejin Varghese and M~Sambath. 2024.
\newblock \href {https://ieeexplore.ieee.org/abstract/document/10533619/} {Yolov8: A novel object detection algorithm with enhanced performance and robustness}.
\newblock In \emph{2024 International Conference on Advances in Data Engineering and Intelligent Computing Systems}, pages 1--6.

\bibitem[{Vaswani et~al.(2017)Vaswani, Shazeer, Parmar, Uszkoreit, Jones, Gomez, Kaiser, and Polosukhin}]{Vaswani2017AttentionIA}
Ashish Vaswani, Noam Shazeer, Niki Parmar, Jakob Uszkoreit, Llion Jones, Aidan~N Gomez, \L~ukasz Kaiser, and Illia Polosukhin. 2017.
\newblock \href {https://proceedings.neurips.cc/paper_files/paper/2017/file/3f5ee243547dee91fbd053c1c4a845aa-Paper.pdf} {Attention is all you need}.
\newblock In \emph{Advances in Neural Information Processing Systems}, volume~30. Curran Associates, Inc.

\bibitem[{Vladymyrov et~al.(2024)Vladymyrov, Von~Oswald, Sandler, and Ge}]{vladymyrov2024linear}
Max Vladymyrov, Johannes Von~Oswald, Mark Sandler, and Rong Ge. 2024.
\newblock \href {https://proceedings.neurips.cc/paper_files/paper/2024/hash/57a3c602f0a1c8980cc5ed07e49d9490-Abstract-Conference.html} {Linear transformers are versatile in-context learners}.
\newblock \emph{Advances in Neural Information Processing Systems}, 37:48784--48809.

\bibitem[{von Oswald et~al.(2023)von Oswald, Niklasson, Randazzo, Sacramento, Mordvintsev, Zhmoginov, and Vladymyrov}]{von2023transformers}
Johannes von Oswald, Eyvind Niklasson, Ettore Randazzo, João Sacramento, Alexander Mordvintsev, Andrey Zhmoginov, and Max Vladymyrov. 2023.
\newblock \href {https://arxiv.org/abs/2212.07677} {Transformers learn in-context by gradient descent}.
\newblock \emph{Preprint}, arXiv:2212.07677.

\bibitem[{Wang et~al.(2021)Wang, Chen, and Zhu}]{wang2021survey}
Xin Wang, Yudong Chen, and Wenwu Zhu. 2021.
\newblock \href {https://ieeexplore.ieee.org/abstract/document/9392296/} {A survey on curriculum learning}.
\newblock \emph{IEEE transactions on pattern analysis and machine intelligence}, 44(9):4555--4576.

\bibitem[{Wei et~al.(2022)Wei, Tay, Bommasani, Raffel, Zoph, Borgeaud, Yogatama, Bosma, Zhou, Metzler, Chi, Hashimoto, Vinyals, Liang, Dean, and Fedus}]{wei2022emergent}
Jason Wei, Yi~Tay, Rishi Bommasani, Colin Raffel, Barret Zoph, Sebastian Borgeaud, Dani Yogatama, Maarten Bosma, Denny Zhou, Donald Metzler, Ed~H. Chi, Tatsunori Hashimoto, Oriol Vinyals, Percy Liang, Jeff Dean, and William Fedus. 2022.
\newblock \href {https://arxiv.org/abs/2206.07682} {Emergent abilities of large language models}.
\newblock \emph{Preprint}, arXiv:2206.07682.

\bibitem[{Wu et~al.(2021)Wu, Dyer, and Neyshabur}]{wu2020curricula}
Xiaoxia Wu, Ethan Dyer, and Behnam Neyshabur. 2021.
\newblock \href {https://arxiv.org/abs/2012.03107} {When do curricula work?}
\newblock \emph{Preprint}, arXiv:2012.03107.

\bibitem[{Xie et~al.(2022)Xie, Raghunathan, Liang, and Ma}]{xie2022an}
Sang~Michael Xie, Aditi Raghunathan, Percy Liang, and Tengyu Ma. 2022.
\newblock \href {https://openreview.net/forum?id=RdJVFCHjUMI} {An explanation of in-context learning as implicit bayesian inference}.
\newblock In \emph{International Conference on Learning Representations}.

\bibitem[{Zhao et~al.(2021)Zhao, Wallace, Feng, Klein, and Singh}]{zhao2021calibrate}
Zihao Zhao, Eric Wallace, Shi Feng, Dan Klein, and Sameer Singh. 2021.
\newblock \href {http://proceedings.mlr.press/v139/zhao21c.html} {Calibrate before use: Improving few-shot performance of language models}.
\newblock In \emph{International conference on machine learning}, pages 12697--12706.

\end{thebibliography}
\newpage
\clearpage
\appendix

\section{Related Work}
\label{chap:context_learning}

\textbf{ICL capabilities of LLMs.} In-context learning ~\cite{brown2020language} is a learning paradigm that models learn intrinsic logical relationships to make accurate predictions without parameter updates. This capability proves particularly advantageous for many NLP tasks, numerical tasks like linear regression \cite{garg2022can,fu2024transformers}, and Markov chain settings \cite{edelman2024evolution,chen2024unveiling,bondaschi2025from}. The early exploration work \cite{elhage2021mathematical,olsson2022context} explains the ICL modeling via the \textit{induction heads}. \citet{lu-etal-2024-emergent} empirically established that LLMs' fine-tuning efficacy stems from ICL. Their large-scale tests in zero-shot settings revealed poor performance across models, confirming ICL as essential for emergent abilities. \citet{grazzi24mamba-icl} extended this research to the Mamba architecture, observing similar ICL capabilities in linear regression tasks. One can find more details in the survey of \citet{dong2024survey} and references therein.

To further study whether the attention mechanisms are the key components in ICL, \citet{ren-etal-2024-identifying} and \citet{todd2024functionvectorslargelanguage} identified distinct yet effective self-attention head types facilitating parameter transfer during ICL. \citet{zhao2021calibrate} demonstrated significant output variations based on prompt ordering, a sensitivity further validated \cite{lu2022fantastically,Chan2022TransformersGD}.

\textbf{ICL for linear tasks.} \citet{NEURIPS2023_b2e63e36} pioneered adaptive algorithm selection via ICL, enabling models to integrate statistical methods for regression problems. \citet{garg2022can} employed a decoder-only GPT-2 Transformer (12 layers, 8 attention heads, and embedding dimension 256) for numerical tasks. Their model processes linear regression coefficients $w_i$ (matrix notation) with randomly sampled $(\vx_k^{(i)}, \vy_k^{(i)})$ pairs, forming ICL sequences $(\vx_1^{(i)}, \vy_1^{(i)}, ..., \vx_N^{(i)}, \vy_N^{(i)}, \vx^{(i)}_\text{query})$. Zero-padding aligns function values with inputs before projection into the embedding space via fully connected layers. Post-Transformer processing maps embeddings back to the output space, demonstrating both in-distribution learning of regression algorithms and out-of-distribution generalization. However, these models are only for linear tasks. But, the generalization ability of ICL is highly nonlinear.

\section{More Experimental Details}
\label{appendix:more-exp-details}

\textbf{Model size and budget.} Consistent with the experimental setup in \cite{garg2022can}, our model comprises either 12 or 24 network layers, equipped with 8 attention heads, and employs a 256-dimensional embedding space. The model was trained on NVIDIA RTX A6000 GPUs, with an average training time of approximately 50 hours for 600k steps.

\textbf{Software packages. }
We employed the \texttt{solve\_ivp} function from the SciPy (Scientific Python) package to compute numerical solutions of differential equations required for training.

\textbf{Training techniques.} \citet{bengio2009curriculum} implemented curriculum learning, progressively increasing problem dimensionality to mirror human learning curves and reduce computational costs \cite{wu2020curricula, wang2021survey}. Cosine annealing \cite{johnson2023ps} optimized training via cyclical learning rate adjustments (peaking then decaying following cosine curves), preventing local optima convergence—a technique widely adopted in models like YOLO-v8 \cite{10533619} and DeepSeek-V3 \cite{DBLP:journals/corr/abs-2412-19437}.

Departing from approach of \citet{garg2022can}, we introduce sliced mean squared error (Sliced-MSE) as the optimization objective. For vector $\vv = (v_i)^n_{i=0}$ and integer $k_v\leq n$, define slice $\vv(:k_v) = (v_0, ..., v_{k_v})$:

\begin{definition}[Sliced Mean Squared Error]
\label{def:sliced_mse}
For ground truth $\vy \in \R^d$ and prediction $\hat{\vy}\in \R^d$, the Sliced-MSE at $\text{Steps}\leq d$ is:
$${\ell}(\vy, \hat{\vy}) = \text{MSE}\left(\vy(:\text{Steps}), \hat{\vy}(:\text{Steps})\right). $$
\end{definition}

We implement progressively complex ODE IVPs, initially testing a 12-layer GPT-2 model before parallel evaluations with a 24-layer variant. 

\begin{figure}[htpb]
    \centering
    \includegraphics[width=0.95\linewidth]{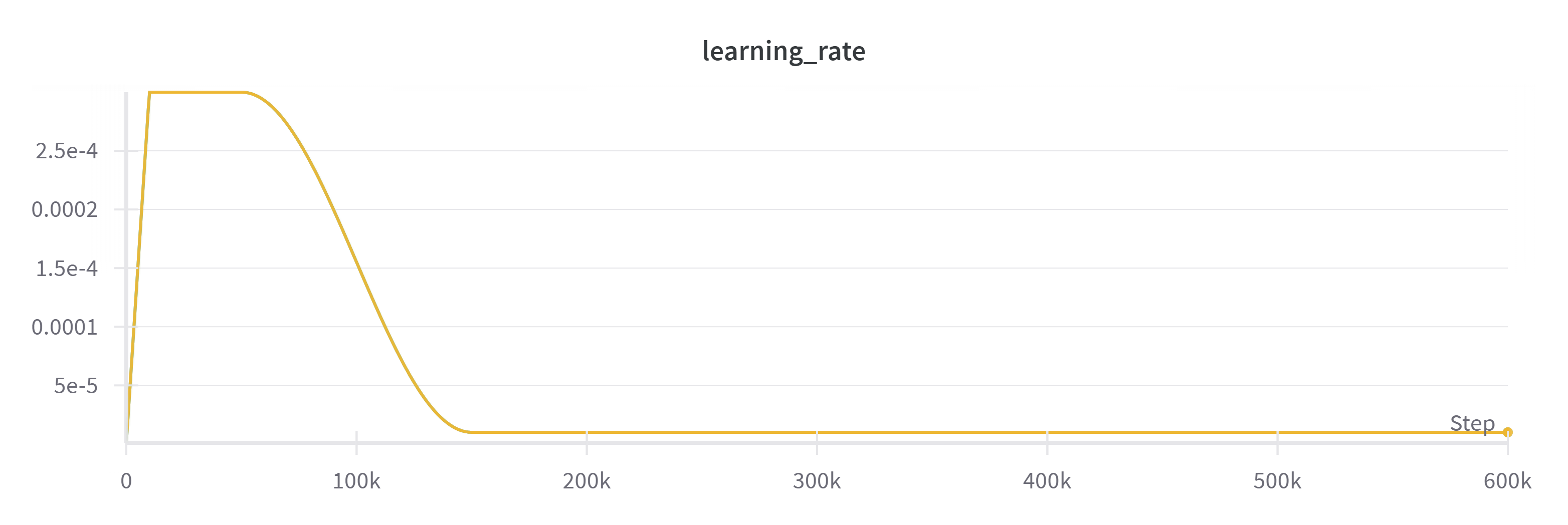}
    \caption{
        Three-phase learning rate schedule combining warm-up, plateau, and cosine annealing. Initial rate $1\times10^{-6}$ linearly increases to $3\times10^{-4}$ over 10k steps, maintains for 40k steps, then decays via cosine annealing to $1\times10^{-5}$ over 10k steps before stabilization.
    }
    \label{fig:learning_rate_setting}
\end{figure}

The training protocol adapts DeepSeek-V3's learning rate scheduling \cite{DBLP:journals/corr/abs-2412-19437} with modifications:
\begin{itemize}
    \item \textbf{Warm-up phase}: Linear learning rate increase with sample size
    \item \textbf{Plateau phase}: Stabilized learning rate period
    \item \textbf{Cosine decay}: Gradual reduction following cosine annealing \cite{johnson2023ps}
\end{itemize}
This multi-stage approach (detailed in Section \ref{sec: further trial}, Figure \ref{fig:learning_rate_setting}) enhances parameter stability during early training while promoting eventual convergence.

Complementing this, we employ curriculum learning to gradually increase problem dimensionality over 30k steps, lowering initial training difficulty and accelerating meaningful parameter acquisition.

While training the first-order ODE, we implement an adjusted cosine annealing schedule (Figure~\ref{fig:learning_rate_setting}).  For specific research questions, it can be found after Definition \ref{def:simple-ivp} and Definition \ref{def:1_order_ivp}

\textbf{IVP formalization.} Here we formally define the initial value problem of our research object as follows:
\begin{definition}[Initial Value Problem]
\label{def:ivp} For mapping $f: \Omega \rightarrow \R$ with open domain $\Omega \subseteq \R \times \R$, an IVP exists given initial condition $(t_0, u_0) \in \Omega$ satisfying:
\begin{align}
    \begin{cases}  
        \frac{\text{d}u}{\text{d}t} = f(u, t), \\  
        u(0) = u_0, \quad  t \in [0, t_e ]. 
    \end{cases}
\end{align} 
\end{definition}
For specific research questions, it can be found in Definition \ref{def:simple-ivp} and Definition \ref{def:1_order_ivp}.

\section{More Experimental Results}

\subsection{Preliminary Exploration: Predictive Accuracy of ICL Models}\label{appendix: first trail}

This section investigates the efficacy of in-context learning (ICL) for nonlinear differential equation solving through a fundamental initial value problem. Our experiments demonstrate that the model successfully predicts solutions within acceptable error margins, with prediction accuracy exhibiting exponential convergence as context length increases. These findings reveal ICL's substantial potential for nonlinear numerical problems when properly trained.

Current research lacks comprehensive exploration of ICL's capabilities for nonlinear numerical solutions. As established earlier, differential equation solving inherently involves nonlinear characteristics. To facilitate the model's initial foray into this domain, we begin with the most elementary form of initial value problems under our framework, which we term the Simple Initial Value Problem (Simple-IVP):

\begin{definition}[Simple Initial Value Problem (Simple-IVP)]
\label{def:simple-ivp} 
A simplified form of Definition \ref{def:ivp} is given by:
\begin{equation} 
\begin{cases}
    f(u, t) = ay + b, \\
    y(0) = y_0, \quad t\in [0, t_e].
\end{cases}
\end{equation}
The input data distribution $\mathcal{D}_{x}= \{\vx_i \colon \vx_i = (\text{Para}_i, t_{e}, \text{Steps}_i)\}$ for Simple-IVP contains parameter sets $\text{Para}_i$ with \textbf{three degrees of freedom} corresponding to coefficients $a, b, y_0$.
\end{definition}

For initial model exploration, we used 12-layer GPT-2 model and  employed a curriculum learning scheme with a fixed learning rate of $10^{-4}$ over 200k training steps. Figure~\ref{fig:warmup_trail} demonstrates the model's performance:

\begin{figure}[H]
    \centering
    \includegraphics[width=\linewidth]{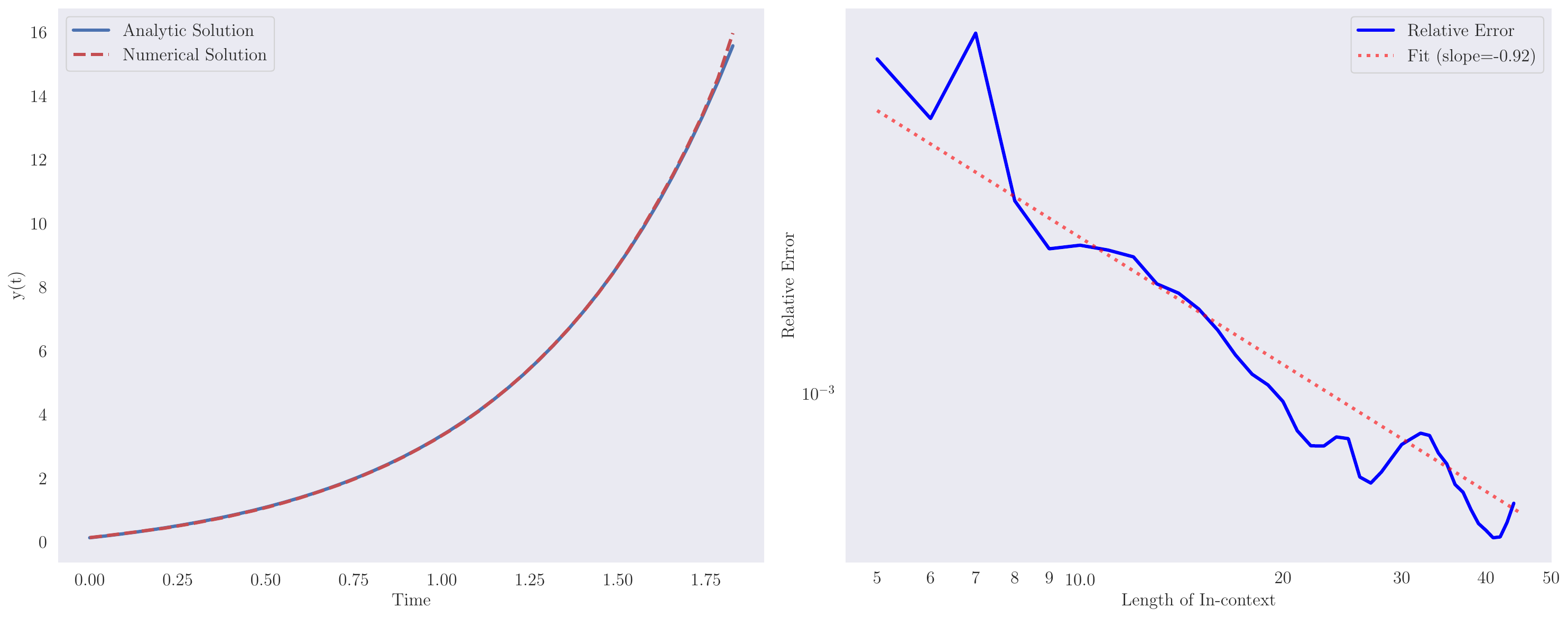}
    \caption{
        Performance of the Transformer model during preliminary training. 
        Left: Solution curve with context length $=40$ for parameters $a=1.7$, $b=1.0$, $y_0=0.1$, $t_e=1.9$, showing near-perfect alignment with ground truth.
        Right: Log-log plot of Sliced-MSE versus context length, with fitted slope $-0.92$ confirming the convergence properties of the ICL approach.
    }
    \label{fig:warmup_trail}
\end{figure}

The model demonstrates competent numerical solving capabilities, with two key observations:
\begin{itemize}
    \item The solution curve exhibits close approximation to the analytical solution
    \item The linear relationship in the log-log error plot reveals exponential convergence of estimation error with increasing context length
\end{itemize}

\subsection{Further training settings}
\begin{definition}[First-Order Linear ODE]
\label{def:1_order_ivp}
A first-order linear ordinary differential equation relates a function to its first derivative through:
\begin{equation}
    \begin{cases}  
        \frac{\text{d}y}{\text{d}t} + p(t)y = q(t), \\  
        y(0) = y_0, t \in [0, t_e], 
    \end{cases}
\end{equation}
where $p(t) = \alpha_1 t + \alpha_2$, $q(t) = \beta_1 e^{\beta_2 t}$. The input distribution $\mathcal{D}_{x}= \{\vx_i \colon \vx_i = \left(\text{Para}_i, t_{e}, \text{Steps}_i\right)\}$ exhibits \textbf{five degrees of freedom} in $\text{Para}_i$: $\alpha_1, \alpha_2, \beta_1, \beta_2, y_0$. 
\end{definition}

\noindent\textbf{Training Configuration:} Both models employ curriculum learning, gradually expanding context length to 41 and vector dimensionality to 64 over the first 30k training steps. The 24-layer variant was introduced for comprehensive comparison alongside the original 12-layer architecture. All models were trained for 600k steps before evaluation.

\subsection{Limitations of In-Context Precision}
\label{appendix: limitations on zero case}
\begin{figure}[htpb]
    \includegraphics[width=0.49\linewidth]{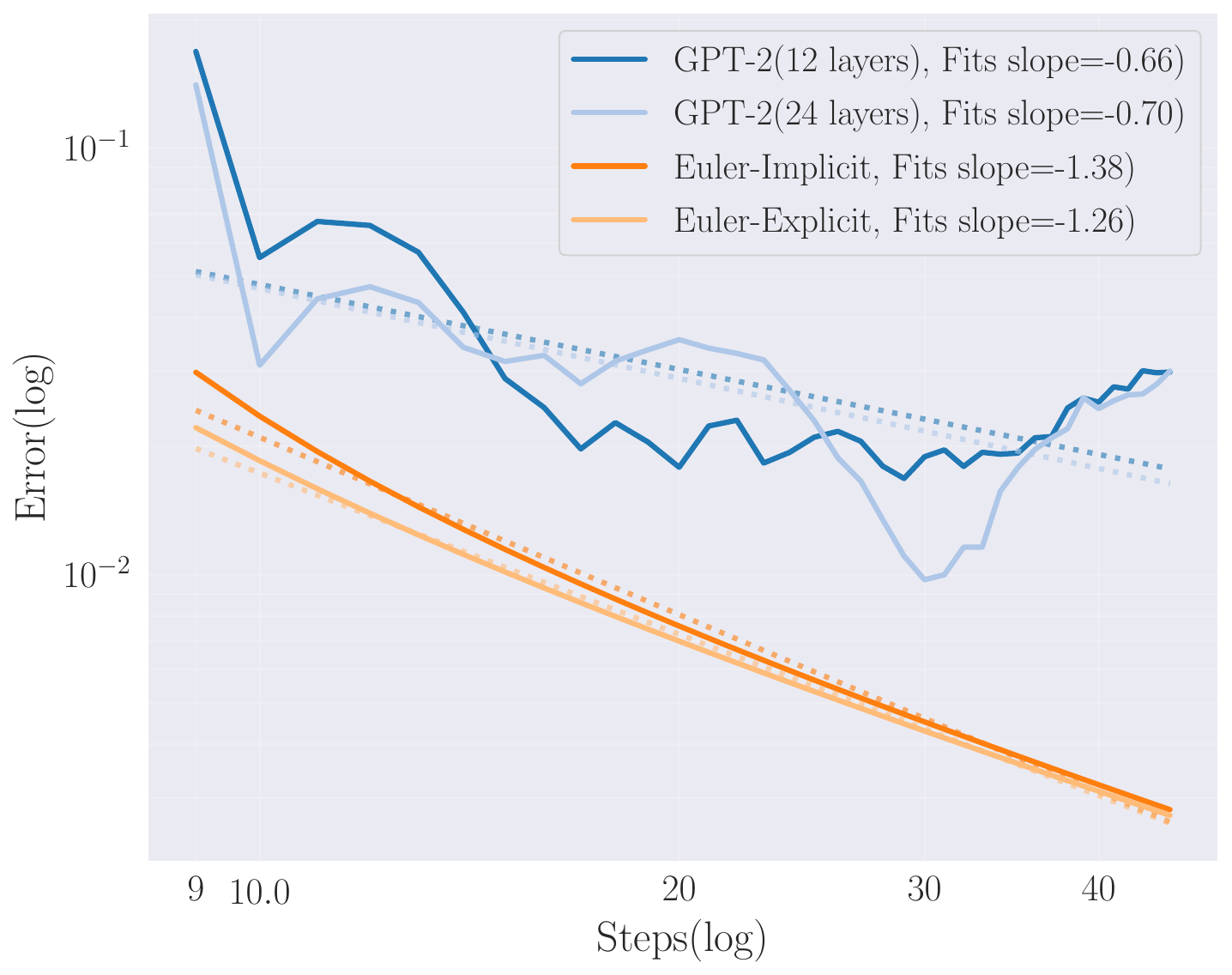}
    \includegraphics[width=0.49\linewidth]{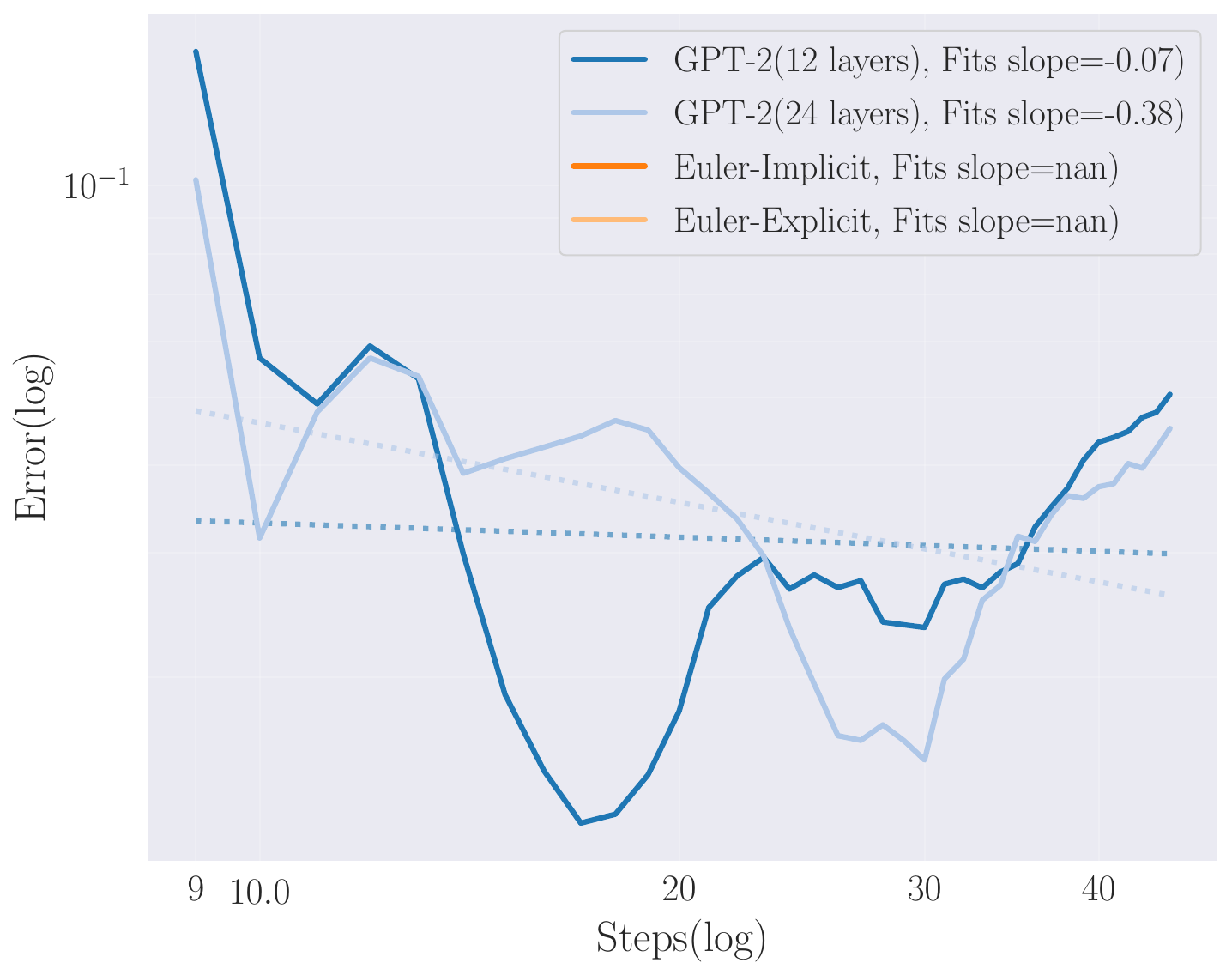}
    \caption{Performance on edge cases (initial value $y(0) = 0.6$). Left: $\alpha_2=1$ with other parameters zero. Right: All parameters zero.}
    \label{fig:nonstable_with_classic}
\end{figure}

Figure \ref{fig:nonstable_with_classic} shows the model's performance under specific parameter conditions. It can be observed that for zero-solution and low-rigidity solutions, the classical solution achieves zero error and low error perfectly, while the model maintains its original accuracy as expected.

\subsection{Another Composite Testing}\label{appendix: more composite}
\begin{figure}[ht!]
      \includegraphics[width=\linewidth]{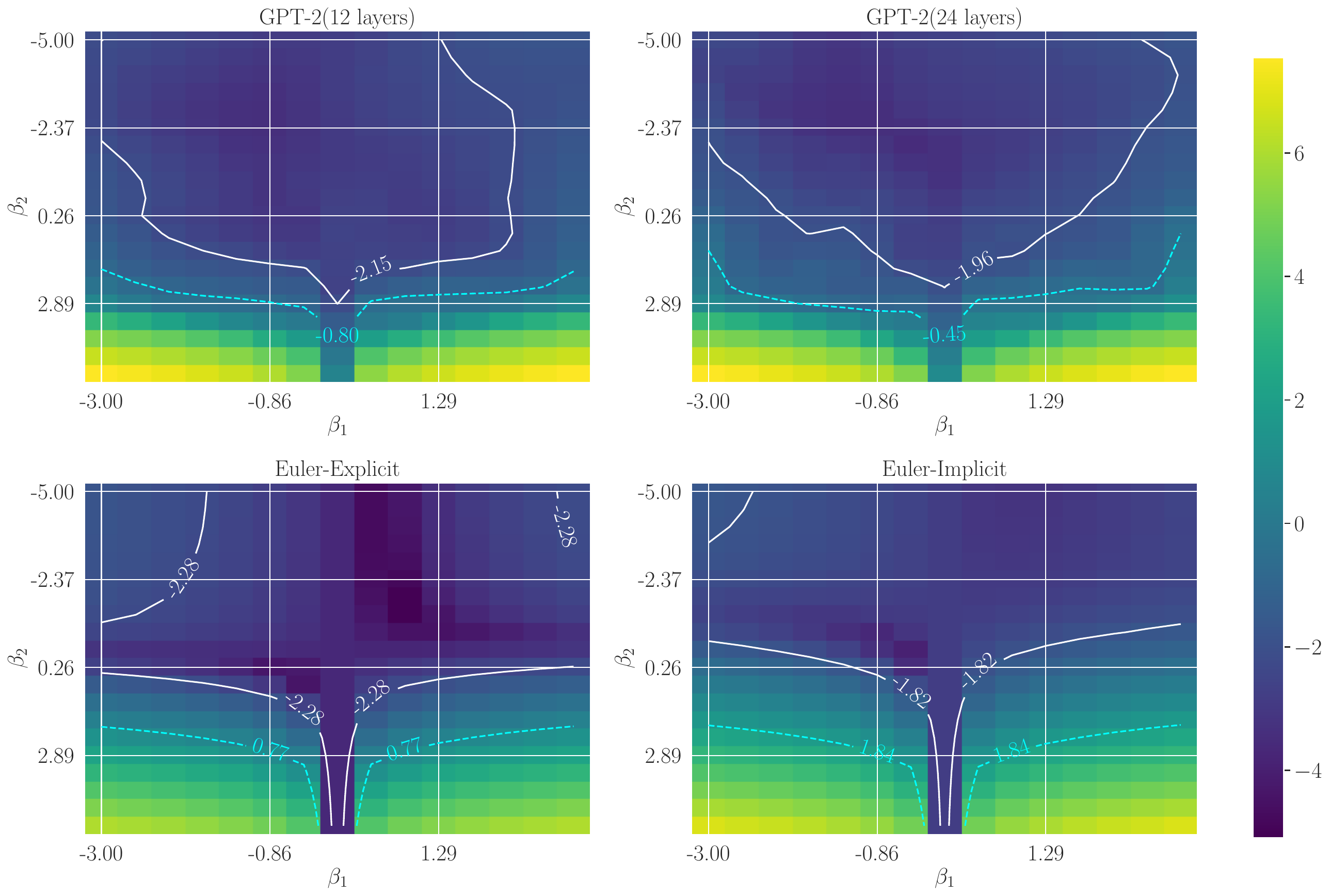}
      \includegraphics[width=\linewidth]{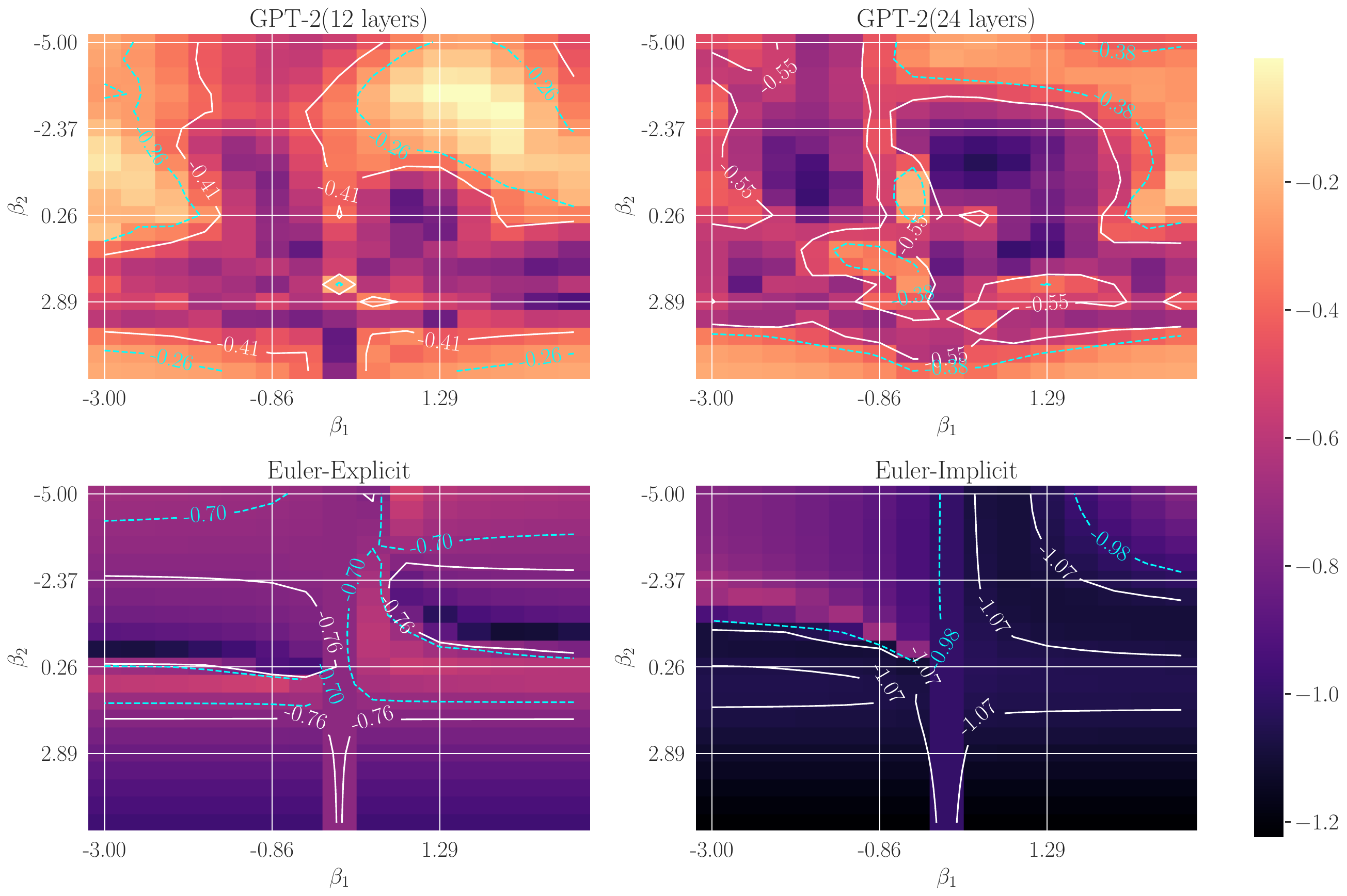}
      \caption{$\beta_1$-$\beta_2$ test region comparisons across parameter combinations.\textit{Upper}: error heatmap; \textit{Lower}: convergence slop (trained on $[-2, 2]\times[-3, 3]$, tested on $[-3, 3]\times[-5, 5]$). In Each subfigure: \textit{Top}: 12L/24L GPT-2;  \textit{Bottom}: Euler Explicit/Implicit. Contours mark 50\% (cyan) and 70\% (white) of each subplot's range.}
      \label{fig:conbined of b_1b_2}
\end{figure}

The patterns observed in the slope heatmap are consistent with the findings in the main text, though with some notable variations. The error heatmap (Fig.~\ref{fig:conbined of b_1b_2}) suggests that GPT-2 solutions tend to maintain broader stable regions in the $\beta$-parameter space, though with relatively modest precision improvements. This observed pattern could potentially relate to the input sequence ordering effect discussed in prior works \cite{zhao2021calibrate}. As $\beta$-parameters typically appear later in the input sequence than $\alpha$-parameters, the self-attention architecture may allocate comparatively less attention weight to these parameters during feature processing. Such positional bias, if present, might simultaneously explain the preserved solution stability (through more consistent global patterns) and the limited precision gains (due to reduced focus on later inputs). However, this interpretation requires further verification as the underlying mechanisms remain incompletely understood.

\section{A Conjecture of In-Context ODE solver}
\label{appendix: conj}

Building upon the observed relationship between convergence accuracy and context length, this study proposes a conjecture (inspired by \citet{liu2025learntooptimizecapabilitiestransformersincontext}.) regarding the convergence properties of in-context learning for ODE solving. We consider this conjecture could provide theoretical foundations for Transformer applications in differential equation solving.

\begin{conjecture}[Convergence of In-Context Learning for ODE Solving]
\label{conj: exponential}
Let $\delta\in(0, 1)$, $c$ be a positive constant. For a Transformer model with $L$ layers and $H$ attention heads, when $N_0 \in [1\colon N-1]$ satisfies specific condition $P(\delta, c, L, H)$, there exists a parameter set such that for any $n \in [N_0\colon N-1]$, the query result $\vy_{n+1}$ of randomly generated first-order linear ODEs and model prediction $\hat{\vy}$ satisfy with probability at least $1-\delta$:
    \begin{equation}
        \label{eq:prop1}
        ||\vy_{n+1} - \hat{\vy}|| \leq ce^{-kN}, \quad N \geq N_0
    \end{equation}
indicating exponential convergence of prediction accuracy with increasing context length.
\end{conjecture}

\end{document}